\documentclass[conference]{IEEEtran}

\makeatother

\IEEEoverridecommandlockouts
\usepackage{placeins}

\usepackage{cite}
\usepackage{amsmath,amssymb,amsfonts}
\usepackage{graphicx}
\usepackage{textcomp}
\usepackage{xcolor}
\usepackage{multirow}
\usepackage{makecell}
\usepackage{subcaption}
\usepackage{caption}
\usepackage{booktabs}
\usepackage{enumitem}
\usepackage{eso-pic}

\def\BibTeX{{\rm B\kern-.05em{\sc i\kern-.025em b}\kern-.08em
    T\kern-.1667em\lower.7ex\hbox{E}\kern-.125emX}}

\begin{document}
\title{\vspace*{1cm} Unified CT–MRI Pancreas Segmentation for Label-Efficient Cross-Modality Subregion Transfer}

\author{
\begin{tabular}{c}
Ziliang Hong$^1$, Hongyi Pan$^1$,
Halil Ertugrul Aktas$^1$, Andrea Bejar$^1$,
Elif Keles$^1$,\\
Frank H. Miller$^1$, Michael B. Wallace$^2$,
Rajesh N. Keswani$^3$, Gorkem Durak$^1$,\\
Alpay Medetalibeyoglu$^4$, Candice Bolan$^2$,
Zongwei Zhou$^5$, Ulas Bagci$^1$
\end{tabular}
\thanks{This work was supported by NIH grants U01-CA268808 and NHLBI R01-HL171376.}
\\[0.5em]
\small
$^1$Department of Radiology, Northwestern University, Chicago, IL, USA\\
\small
$^2$Division of Gastroenterology and Hepatology, Mayo Clinic Florida,
Jacksonville, FL, USA\\
\small
$^3$Department of Gastroenterology and Hepatology,
Northwestern University, Chicago, IL, USA\\
\small
$^4$Istanbul University, Istanbul, Turkey\\
\small
$^5$Johns Hopkins University, Baltimore, MD, USA
}

\maketitle

\begin{abstract}
Robust medical image segmentation across imaging modalities is challenging because of large differences in appearance and intensity distributions. Models trained on a single modality often show substantial performance drops when applied to unseen domains. In this work, we develop a unified 3D pancreas segmentation framework that applies domain-adversarial learning to 4,604 heterogeneous CT and MRI scans to learn anatomical representations. A shared nnU-Net encoder-decoder is trained for whole-pancreas segmentation, with a latent domain discriminator encouraging CT-MRI feature alignment. The learned encoder is subsequently transferred to pancreatic head-body-tail segmentation using limited MRI-only subregion annotations. An average Dice score of 87.31\% on the in-distribution test set and Dice scores ranging from 84.20\% to 88.09\% across external OOD datasets were achieved in whole pancreas segmentation. Dice scores of 80.53\% on MRI and 83.05\% on CT were achieved for downstream subregion segmentation, without using CT subregion annotations. These results demonstrate that a unified anatomical representation can support both cross-modality pancreas segmentation and label-efficient downstream transfer.
\end{abstract}


\section{Introduction}
\label{sec:intro}
Accurate pancreas segmentation is essential for many clinical and research applications, including disease diagnosis, surgical planning, and downstream analysis~\cite{seyithanoglu2024advances,zhou2017fixed}. However, this task remains challenging due to large anatomical variability, irregular organ shape, and low contrast with surrounding tissues. These difficulties are further amplified across imaging modalities such as CT and MRI, where appearance and intensity distributions differ substantially.

In clinical practice, abdominal imaging is routinely acquired using multiple modalities and MRI sequences (T1-weighted (T1W), T2-weighted (T2W), and out-of-phase (OOP)). Training separate models for each modality is inefficient and often infeasible due to limited annotations. Although a unified multi-modal model is desirable, directly combining CT and MRI data typically leads to performance degradation caused by strong modality-induced domain shifts, resulting in models that rely on modality-specific appearance cues rather than stable anatomical representations.

To address this challenge, learning modality-invariant anatomical features is critical for robust cross-modality pancreas segmentation and for downstream tasks such as pancreas subregion segmentation or following analysis~\cite{zhang2025large}. While domain adversarial learning has been explored for cross-modality segmentation, its effectiveness in learning generalizable representations under limited-data or single-modality supervision remains insufficiently studied. Our contributions are summarized as follows:

\begin{itemize}
\item We develop a unified CT--MRI pancreas segmentation framework that uses domain-adversarial learning to learn modality-invariant pancreatic representations, enabling a single model to segment the pancreas across multiple imaging modalities.

\item We demonstrate that the learned representations are transferable to downstream tasks, allowing limited annotations from a single modality to support cross-modality downstream segmentation.

\item Specifically, we achieve pancreatic head--body--tail segmentation on CT without using any CT subregion annotations during downstream training, addressing a clinically relevant setting in which fine-grained anatomical labels are unavailable in the target modality.
\end{itemize}
Code and weights will be provided upon acceptance.
\begin{figure*}[!t]
    \centering
    \includegraphics[width=1\linewidth]{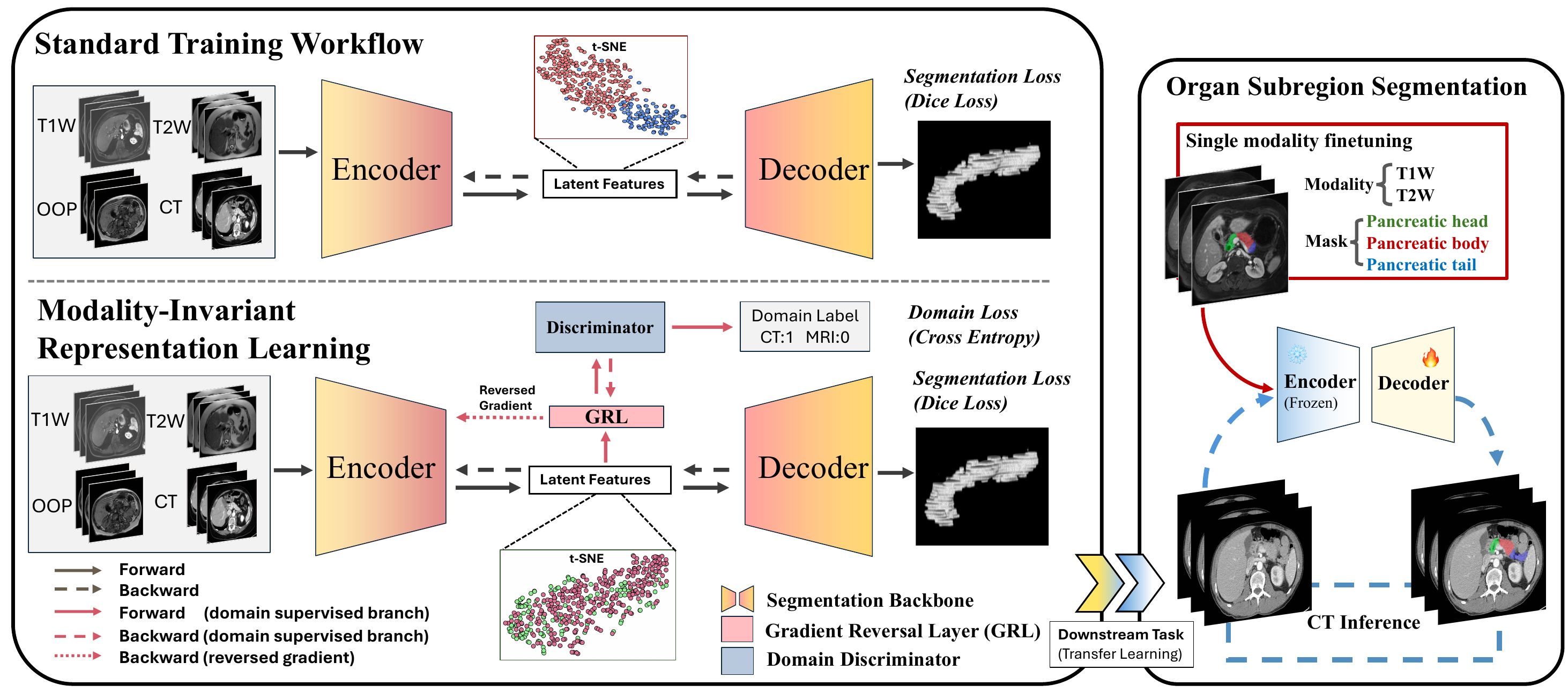}
    \caption{Modality-invariant representation learning. Mixed CT and MRI images are input to a shared encoder-decoder. Latent features pass through a GRL into a domain discriminator for modality prediction, enforcing modality-invariant representations. The decoder is trained with standard segmentation loss, and the learned encoder is transferred to pancreas subregion segmentation under limited-data or single-modality fine-tuning.
    }
    \label{workflow}
\end{figure*}

\section{Related Work}
\subsection{Domain Adversarial Learning}
Domain adversarial learning has been widely studied to mitigate domain shift by encouraging domain-invariant feature representations. A seminal work in this area is the Domain-Adversarial Neural Network~\cite{ganin2016domain,sicilia2023domain,zhao2018adversarial}, which introduces a gradient reversal layer to align feature distributions between source and target domains through adversarial training. This framework has inspired many extensions in both computer vision and medical image analysis.

In medical imaging, domain adversarial learning has been widely adopted to address variations introduced by different imaging protocols, scanners, and acquisition modalities. Kamnitsas et al.~\cite{kamnitsas2017unsupervised} employed adversarial training for unsupervised domain adaptation in brain lesion segmentation, achieving improved generalization across datasets. Beyond cross-dataset settings, adversarial learning has also been extended to cross-modality scenarios, where aligning latent feature representations between CT and MRI has been shown to effectively reduce modality discrepancies in organ segmentation~\cite{chen2019synergistic,guan2021domain}.

\subsection{Pancreas Segmentation}
Automatic pancreas segmentation has long been recognized as a challenging task due to the organ's large anatomical variability, low contrast, and irregular shape. Early methods relied on multi-atlas registration and handcrafted features~\cite{zhou2017fixed}, but were later surpassed by deep learning-based approaches.

With the success of convolutional neural networks, architectures such as U-Net~\cite{ronneberger2015u} and its variants have become the mainstream for organ segmentation~\cite{roth2015deeporgan,pan2025adaptive}. nnU-Net later provided a standardized and robust framework that adapts network configurations to specific datasets, achieving strong performance across a wide range of medical segmentation tasks~\cite{isensee2021nnu}.

Despite these advances, most pancreas segmentation methods are still designed for CT. Pancreas segmentation in MRI remains relatively underexplored, largely due to the limited availability of annotated datasets and substantial appearance differences across T1W and T2W sequences. Recently, PaNSegNet~\cite{zhang2025large,proietto2021hierarchical} demonstrated state-of-the-art performance using large-scale MRI data, pushing this challenge to the next level: learning representations that are robust and transferable across imaging modalities, which has not been fully addressed.

\section{Methodology}
\subsection{Problem Statement}
Let $\{\mathbf{X}, \mathbf{Y}\}$ be a multi-modal pancreas segmentation dataset containing $N$ images. For the $i$-th sample, $\mathbf{x}_i \in \mathbf{X}$ denotes the input image, $\mathbf{y}_i \in \mathbf{Y}$ represents the ground-truth segmentation mask, and $d_i \in \{0, 1\}$ indicates the modality label (MRI: 0; CT: 1). The standard segmentation network, such as nnUNet~\cite{isensee2021nnu}, contains an encoder $\mathcal{E}$ and a decoder $\mathcal{D}$. It generates a segmentation prediction for the given image: $\hat{\mathbf{y}}_i = \mathcal{D}(\mathcal{E}(\mathbf{x}_i))$. The primary segmentation objective is to minimize a joint loss function comprising the Dice loss and Cross-Entropy (CE) loss between the segmentation prediction and the ground-truth segmentation mask to capture both global overlap and voxel-wise accuracy:
\begin{equation}
\mathcal{L}\mathrm{seg} = \sum_{i=0}^{N-1} \left( \mathcal{L}_{\mathrm{Dice}}(\hat{\mathbf{y}}_i, \mathbf{y}_i) + \mathcal{L}_{\mathrm{CE}}(\hat{\mathbf{y}}_i, \mathbf{y}_i) \right).
\end{equation}
While effective for a single modality, the significant domain shift between MRI and CT often degrades the generalizability of the model to unseen distributions.

\subsection{Modality-Invariant Representation Learning}
To mitigate domain shift, we adopt a modality-invariant representation learning strategy based on the standard domain-adversarial neural network (DANN)~\cite{ganin2016domain}, introducing a domain discriminator $\mathcal{C}$ as illustrated in Fig.~\ref{workflow}. The domain discriminator $\mathcal{C}$ provides an adversarial supervisory signal to the encoder $\mathcal{E}$, effectively penalizing the extraction of modality-specific features. By competing against the discriminator, the encoder is encouraged to map both CT and MRI inputs into a latent space where anatomical representations are domain-invariant and robust across modalities.

\subsubsection{Domain Discriminator and Joint Optimization}
The domain discriminator is trained by minimizing the cross-entropy loss:
\begin{equation}
\mathcal{L}_{\mathrm{domain}}
=
\sum_{i=0}^{N-1}\mathcal{L}_\mathrm{CE}\left(\mathcal{C}(\mathcal{E}(\mathbf{x}_i)),\, d_i\right).\label{eq:domain loss}
\end{equation}
This component aims to identify the imaging modality of the input based on the latent features produced by the encoder. By minimizing this loss, the discriminator can distinguish between MRI and CT  distributions.
To stabilize domain adversarial training, we follow~\cite{ganin2016domain} to insert a gradient reversal layer (GRL) $\mathcal{R}$
between the segmentation encoder and the domain discriminator.
The GRL acts as an identity transformation during the forward pass but reverses the gradient during backpropagation:
\begin{equation}
\mathcal{R}(\mathbf{x}) = \mathbf{x}, \quad \frac{d\mathcal{R}}{d\mathbf{x}} = -\mathbf{I},
\end{equation}
where $\mathbf{I}$ is the identity matrix. This reversal forces the encoder to suppress domain-specific information, thereby learning features that are indistinguishable across modalities. 
The total objective function is formulated as:
\begin{equation}
\mathcal{L}_{\mathrm{total}} = (1-\lambda)\mathcal{L}_{\mathrm{seg}} + \lambda \mathcal{L}_{\mathrm{domain}},
\end{equation}
where $\lambda \in [0, 1]$ is a hyperparameter scaling the influence of the adversarial signal. To ensure the encoder learns meaningful anatomical features and to maintain training stability, we utilize a warm-up strategy: $\lambda$ is initialized at $0$ and gradually increased to a maximum of $0.5$ as training progresses.

\subsubsection{Pancreas Subregion Segmentation}
With preserving modality-invariant representation, the encoder is subsequently reused for the downstream pancreas subregion segmentation task. Given that subregion annotations are available only for a limited subset of data, the encoder is frozen and transferred as a shared feature extractor, while a task-specific decoder is trained using single-modality subregion labels. This design allows the model to benefit from the modality-invariant anatomical representations learned from the upstream pancreas segmentation task. With a robust encoder, the downstream decoder can therefore be adapted using limited subregion annotations from a single modality.

\section{Experiments}
\begin{table}[tb]
    \caption{Dataset distribution. ID and OOD denote in-distribution and out-of-distribution, respectively.}
    \centering
\setlength{\tabcolsep}{3pt}
    \begin{tabular}{l|lcc}
    \toprule
         &\textbf{Dataset}&\textbf{Modality}&\textbf{Scans}\\
         \midrule
         \multirow{4}{*}{ID}&Cyst-X~\cite{pan2025cyst, hong2025pancreas}&MRI&1461\\
         &Private&MRI&1888\\
         &AbdomenCT-1K~\cite{ma2021abdomenct}&CT&1000\\
         &Peri-Pancreatic Edema~\cite{PeripancraticEdema}&CT&255\\
         \midrule
        \multirow{4}{*}{OOD}&AMOS~\cite{ji2022amos}&MRI&60\\
        &U-Mamba~\cite{U-Mamba}&MRI&50 \\
        &AMOS~\cite{ji2022amos}&CT&300\\
        &BTCV~\cite{landman2015miccai}&CT&30\\
         \bottomrule
    \end{tabular}
    \label{tab:dataset distribution}
\end{table}

\subsection{Tasks and Datasets}
We define two pancreatic segmentation tasks to evaluate the efficacy of our framework:

\noindent\textbf{Whole Pancreas Segmentation (Primary):} The primary objective is the binary segmentation of the entire pancreatic volume across CT and MRI modalities. This task is used to train the domain-invariant encoder and is evaluated on both in-distribution (ID) test sets and out-of-distribution (OOD) datasets to assess cross-center and cross-modality robustness.

\noindent\textbf{Anatomical Subregion Segmentation (Downstream):} To assess the representational quality of the learned features, we perform a fine-grained segmentation task. This involves partitioning the organ into three anatomical subregions: the head, body, and tail. This task follows a transfer learning protocol where the domain-invariant encoder is frozen, and a task-specific decoder is trained to delineate these complex structures. Only the training and validation subset from the Cyst-X dataset~\cite{pan2025cyst, hong2025pancreas} is utilized in the fine-tuning. Subregion segmentation of MRI and CT will be evaluated on the Cyst-X test subset and CT images from AMOS~\cite{ji2022amos} and BTCV~\cite{landman2015miccai}.

As summarized in Table~\ref{tab:dataset distribution}, this study utilizes a large-scale collection of pancreas abdominal scans organized into two primary categories:

\noindent\textbf{In-Distribution (ID) Training and Testing:} This cohort comprises 4,604 scans across MRI (Cyst-X~\cite{pan2025cyst, hong2025pancreas} and a private MRI dataset) and CT (AbdomenCT-1K~\cite{ma2021abdomenct} and Peri-Pancreatic Edema~\cite{PeripancraticEdema}) modalities. The ID dataset was partitioned into training, validation, and test sets using an 8:1:1 ratio. Among these, only the Cyst-X dataset provides ground-truth masks for both the whole pancreas and its subregions (head, body, and tail).

\noindent\textbf{Out-of-Distribution (OOD) Evaluation:} To assess generalization under distribution shifts, we evaluated the framework on three external datasets: AMOS~\cite{ji2022amos}, BTCV~\cite{landman2015miccai}, and U-Mamba~\cite{U-Mamba}. Because these datasets lacked subregion annotations, an expert radiologist manually segmented the pancreatic head, body, and tail for a subset of the OOD CT data (7 scan from AMOS and 10 scans from BTCV) to facilitate a rigorous performance validation on the CT modality.

\subsection{Implementation Details}
Experiments were implemented with a standard nnU-Net backbone \cite{isensee2021nnu} and executed on a server with 8 NVIDIA A6000 GPUs. The standard nnU-Net preprocessing pipeline was adopted, with all CT and MRI images and corresponding annotations resampled to the median training-set spacing along each spatial axis. Image intensities were normalized using z-score normalization. The network was trained using 3D patches of $80 \times 160 \times 192$ voxels, and sliding-window inference was used to make whole-volume predictions.
The segmentation network and discriminator were jointly optimized using the Adam optimizer \cite{kingma2014adam} with an initial learning rate of $0.001$. For downstream applications, the domain-invariant encoder is frozen, and the task-specific segmentation decoder is trained independently. We adopted a multi-stage training strategy to balance segmentation accuracy with domain invariance:

\noindent\textbf{Stage I Warm-up:}
The network was trained on combined CT and MRI data for 2,000 epochs using only the segmentation loss $\mathcal{L}_{\mathrm{seg}}$ to stabilize the model's capture of complex pancreatic anatomy before introducing adversarial training.
\begin{figure*}[!t]
    \centering
    \includegraphics[width=\linewidth]{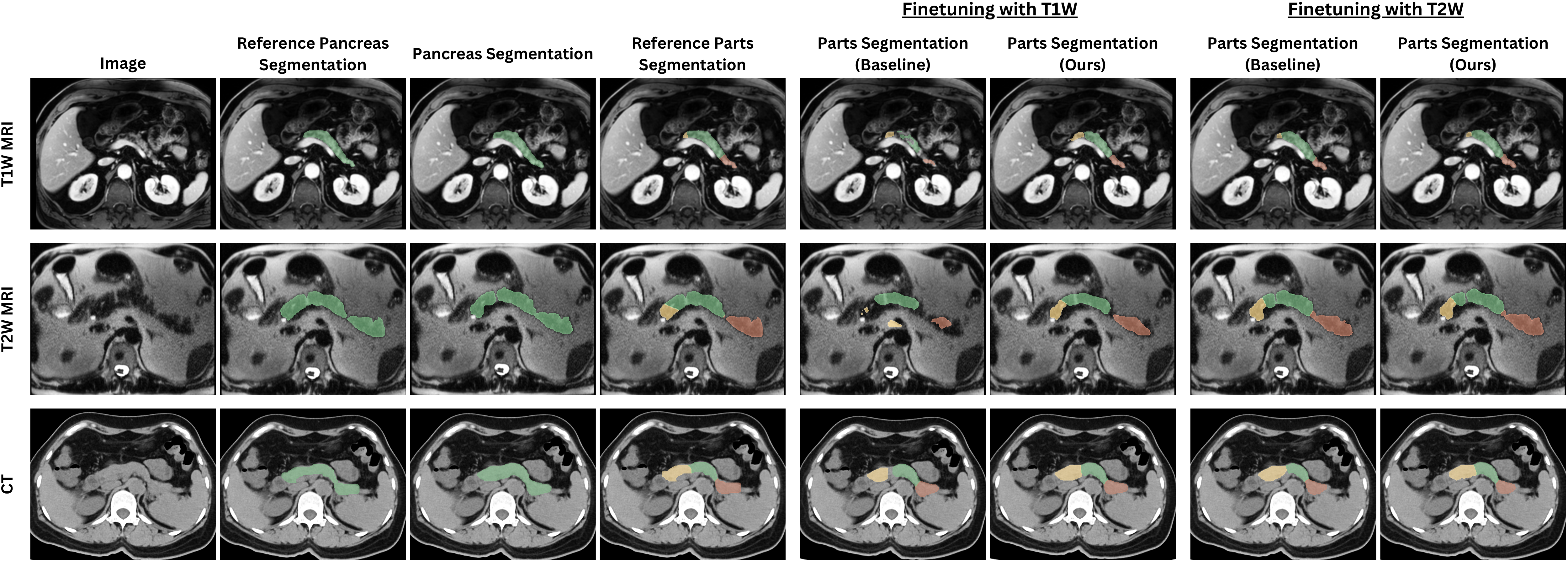}
    \caption{Qualitative visualization of whole-pancreas and subregion segmentation across modalities. In whole pancreas segmentation, the pancreas is shown in green. For pancreas subregion segmentation, yellow, green, and red denote the pancreatic head, body, and tail, respectively. Results are shown for MRI (T1W and T2W) and CT images, comparing the baseline and the proposed method under different fine-tuning settings.}
    \label{fig:segmentation_images}
\end{figure*}

\noindent\textbf{Stage II Adversarial Alignment:} Training continued for 1,000 epochs with the domain discriminator integrated. Domain supervision was applied at the modality level (CT vs. MRI) rather than differentiating between specific MRI sequences. This design is supported by the shared underlying physics of MR imaging and our preliminary t-SNE visualization (Fig. \ref{fig:tsne}), which reveals latent feature distributions across MRI sequences are relatively congruent, while the divergence between CT and MRI remains the main source of domain shift.

\subsection{Experimental Results}
\subsubsection{Whole Pancreas Segmentation}
Fig.~\ref{fig:segmentation_images} shows the whole pancreas segmentation and pancreas subregion segmentation results visualized in 3D Slicer.
Table~\ref{pancreas_seg} summarizes the performance of the proposed model on the in-distribution (ID) test set and multiple out-of-distribution (OOD) datasets. On the ID test set, our framework demonstrates robust and consistent performance across modalities, achieving an overall average Dice score of 87.31\%. Within the MRI sub-modalities, T2W images yield superior boundary accuracy, characterized by the lowest distance-based errors (HD95: 2.923 mm, ASSD: 0.537 mm), indicating highly precise contour delineation. The OOP MRI subset achieves the highest overlap metrics, with a Dice score of 89.76\% and an IoU of 82.67\%. Performance on CT images remains comparable (Dice: 87.53\%), confirming the efficacy of our cross-modality feature alignment.

The model exhibits stable generalization on OOD datasets without the need for additional fine-tuning. Dice scores consistently exceed 84\% across all external benchmarks, highlighting strong robustness to distribution shifts. Specifically, the model achieves Dice scores of 84.20\% on AMOS (CT) and 84.58\% on BTCV (CT). Performance on MRI-based OOD datasets is notably high, reaching 86.87\% on AMOS (MRI) and 88.09\% on the U-Mamba dataset. These results may suggest that the domain-invariant representations generalize effectively across varied scanner protocols and imaging centers.

\begin{table}[tb]
\centering
\caption{Pancreas segmentation performance on 10\% test set and out-of-distribution dataset.}
\label{pancreas_seg}
\setlength{\tabcolsep}{3pt}
\begin{tabular}{llcccc}
\toprule
&\textbf{Dataset} & \textbf{Dice(\%)} & \textbf{HD95} & \textbf{ASSD} & \textbf{IoU(\%)} \\
\midrule
\multirow{4}{*}{Test}
&Overall & 87.31 & 5.229 & 0.940 & 78.42 \\
&T1W & 85.59 & 5.432 & 0.966 & 75.88 \\
&T2W & 88.27 & 2.923 & 0.537 & 80.06 \\
&OOP & 89.76 & 8.469 & 0.827 & 82.67 \\
&CT  & 87.53 & 6.176 & 1.312 & 78.33 \\
\midrule
\multirow{4}{*}{OOD}
&AMOS(CT)         & 84.20 & 7.928 & 0.854 & 74.29  \\
&AMOS(MRI)        & 86.87 & 3.436 & 0.778 & 77.79  \\
&BTCV(CT)         & 84.58 & 3.901 & 0.652 & 73.64  \\
&U-Mamba(MRI)      & 88.09 & 5.814 & 0.992 & 79.34 \\
\bottomrule
\end{tabular}
\end{table}

\begin{table}[tb]
\centering
\caption{Segmentation performance of pancreatic anatomical subregions on MRI and CT test sets.}
\label{tab:pancreas_subregion_seg}
\setlength{\tabcolsep}{3pt}
\begin{tabular}{lccccc}
\toprule
\textbf{Modality} & \textbf{Subregion} & \textbf{Dice(\%)} & \textbf{HD95} & \textbf{ASSD} & \textbf{IoU(\%)} \\
\midrule
\multirow{4}{*}{MRI}
 & Avg & 80.53 & 4.878 & 1.014 & 68.97 \\
 & Head & 85.19 & 3.402 & 0.576 & 74.87 \\
 & Body & 80.23 & 5.289 & 0.769 & 67.84 \\
 & Tail & 76.18 & 5.944 & 1.698 & 64.22 \\
\midrule
\multirow{4}{*}{CT}
 & Avg  & 83.05 & 4.290 & 0.649 & 71.33 \\
 & Head & 82.29 & 4.901 & 0.746 & 70.14 \\
 & Body & 83.26 & 4.613 & 0.464 & 71.59 \\
 & Tail & 83.61 & 3.355 & 0.737 & 72.25 \\
\bottomrule
\end{tabular}
\end{table}

\subsubsection{Pancreas Subregion Segmentation}
The results for the downstream subregion segmentation task are detailed in Table~\ref{tab:pancreas_subregion_seg}. The transferred encoder achieves an average Dice of 80.53\% on MRI and 83.05\% on CT. This result is notable because no CT subregion annotations were used during downstream training

Consistent with anatomical challenges, the pancreatic head demonstrates the highest segmentation accuracy (e.g., 85.19\% in MRI), while the tail remains the most difficult region to delineate (e.g., 76.18\% in MRI) due to its smaller volume and high anatomical variability. However, the remarkably low distance-based errors on CT (Average HD95: 4.290 mm) suggest that the learned representations capture stable anatomical structures that translate well to fine-grained segmentation tasks.

\begin{figure}[!t]
    \centering
    \begin{subfigure}{0.75\linewidth}
        \centering
        \includegraphics[width=\linewidth]{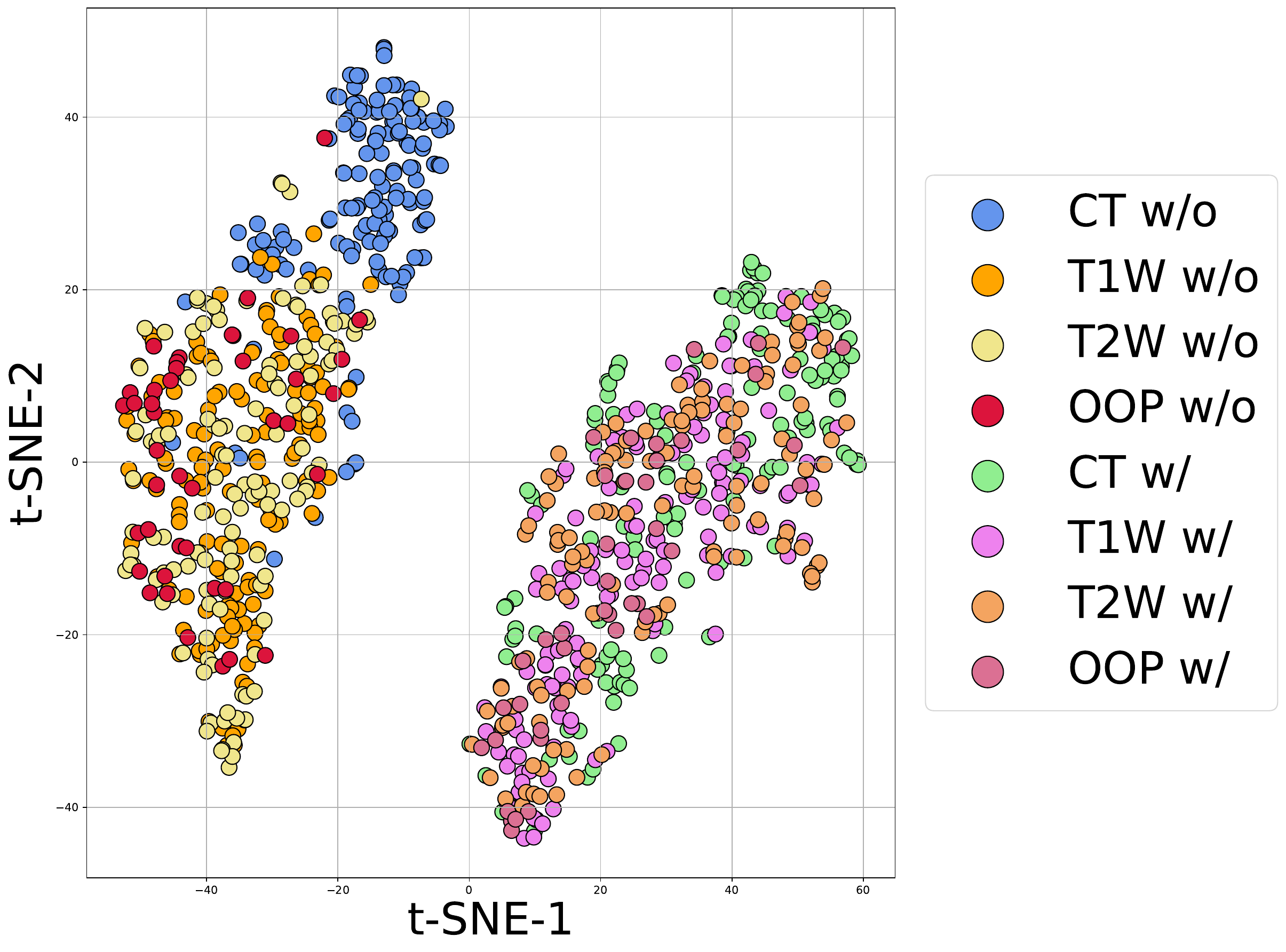}
        \caption{All modalities.}
        \label{fig:tsne_all_modal}
    \end{subfigure}
    \\
    \begin{subfigure}{0.75\linewidth}
        \centering
        \includegraphics[width=\linewidth]{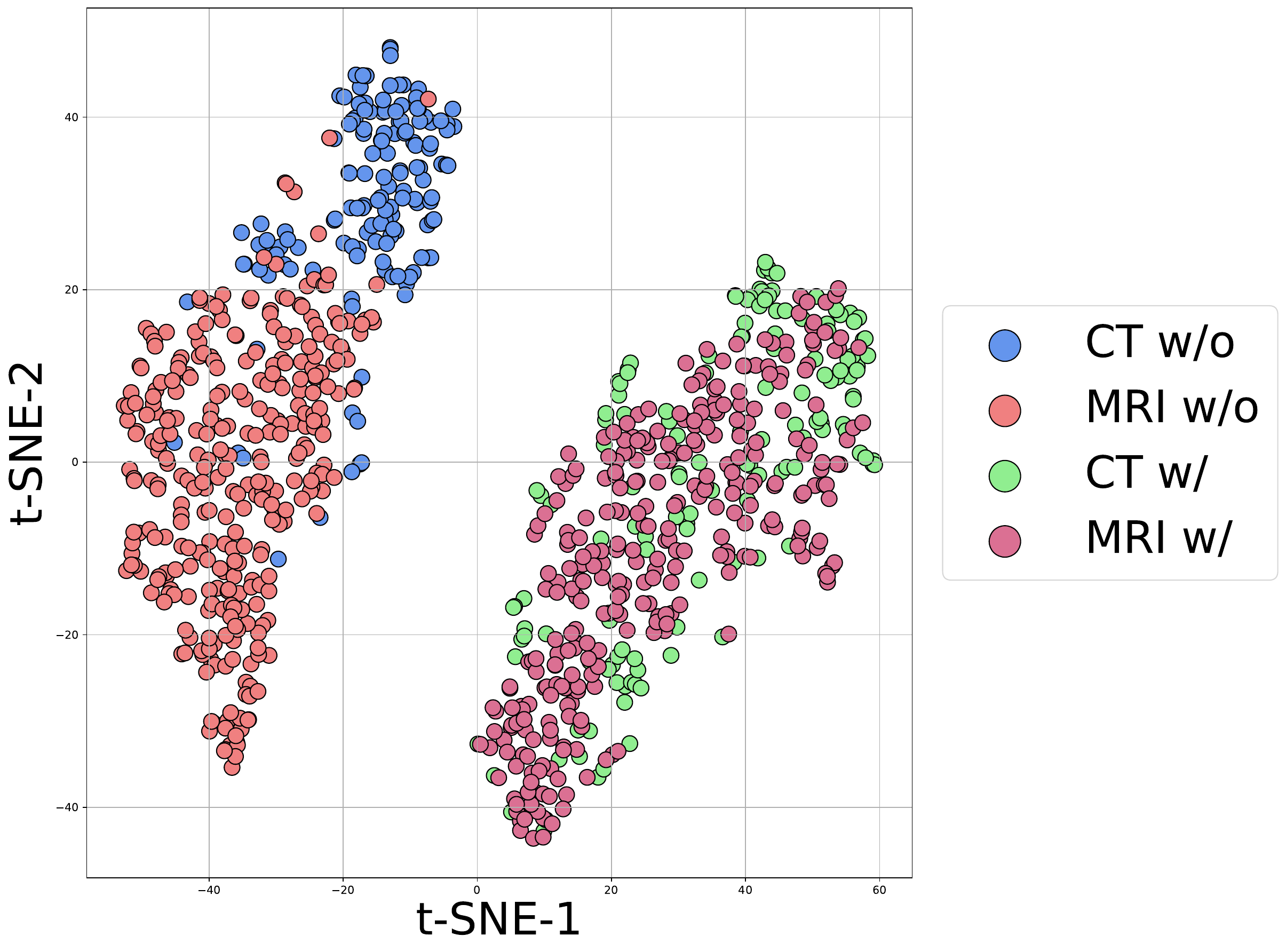}
        \caption{CT vs MRI.}
        \label{fig:tsne_ct_mri}
    \end{subfigure}
    \caption{Visualization of bottleneck features with t-SNE on the test set. (a) MRI sequences and CT are shown in different colors. (b) All MRI sequences are grouped as a single class.}
    \label{fig:tsne}
\end{figure}

\begin{figure}[!t]
    \centering
    \subfloat[MRI dice score.\label{ablation_mri}]{
\includegraphics[width=0.75\linewidth]{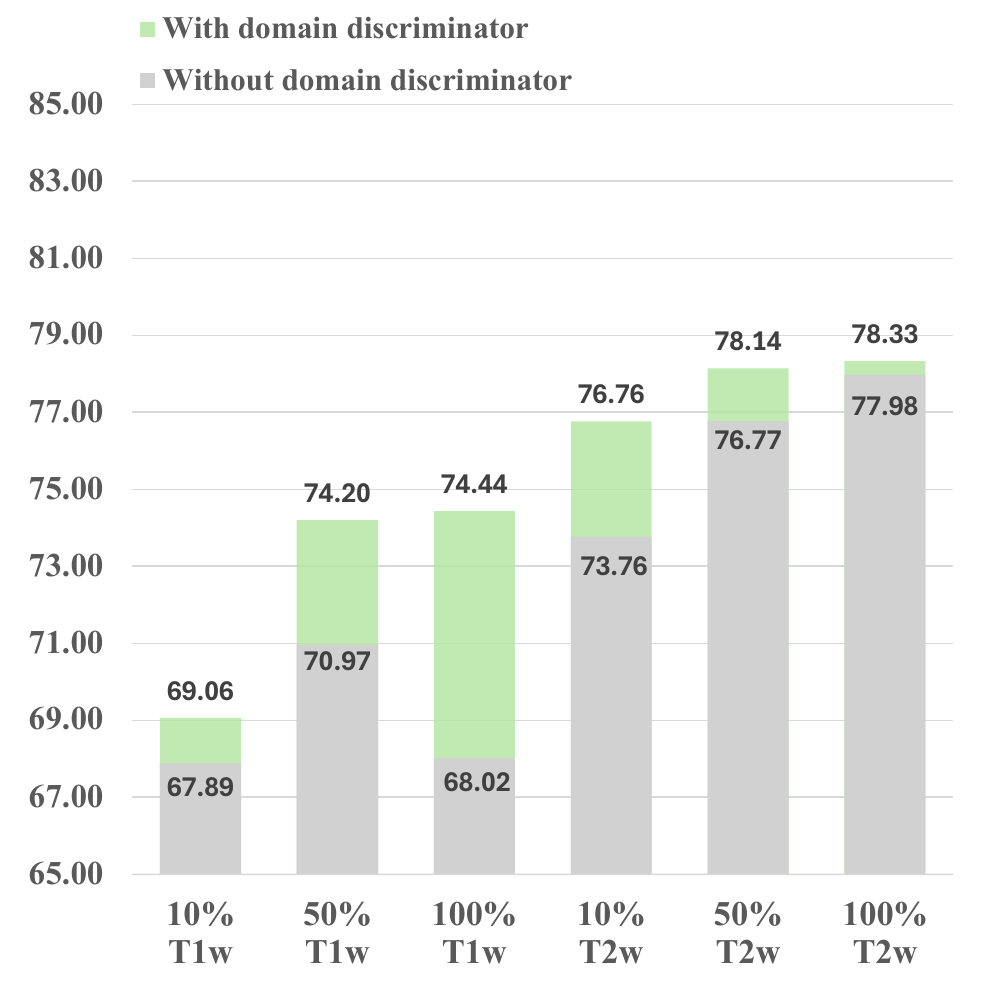}}\\
    \subfloat[CT dice score.\label{ablation_ct}]{\includegraphics[width=0.75\linewidth]{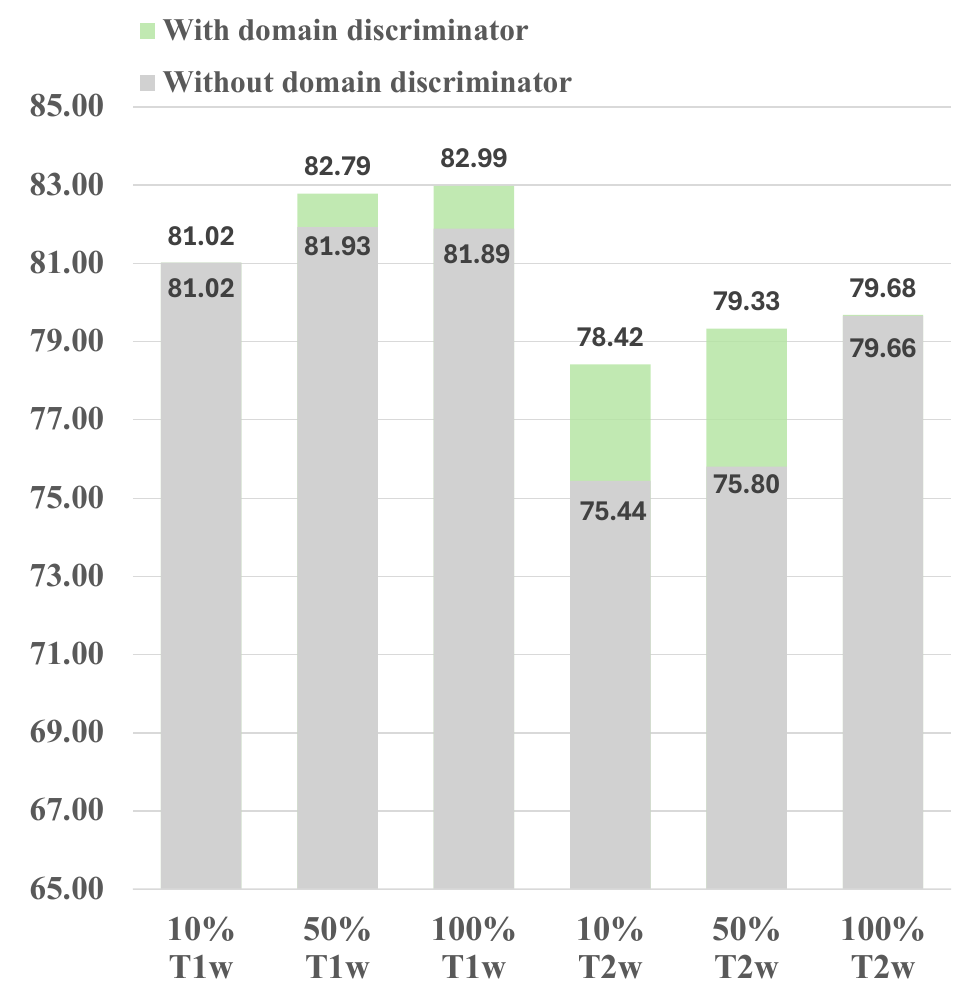}}
    \caption{Partial single-modality fine-tuning. Our modality-invariant representation learning (green) outperforms or is comparable to the baseline (gray) across all settings.}\label{ablation.}
\end{figure}

\subsubsection{Ablation study of partial single-modality fine-tuning} Fig.~\ref{ablation.} presents the ablation study evaluating the effect of domain adversarial training under partial single-modality fine-tuning. The segmentation model is fine-tuned using varying proportions (10\%, 50\%, and 100\%) of MRI data from either T1W or T2W sequences, and evaluated on both MRI and CT test sets.
On the MRI test set, the model with domain-adversarial pretraining outperformed or matched the baseline across all fine-tuning settings, with the largest numerical improvement observed after fine-tuning on 100\% of the T1W data. On the CT test set, the gains were most evident with 10\% and 50\% of the T2W data, whereas several T1W and full-data T2W settings showed only small differences. Overall, these results suggest that the benefit of modality-invariant pretraining depends on both the MRI sequence and the amount of downstream supervision.

\subsection{Discussion}
To further examine the impact of joint optimization, we visualize bottleneck features using t-SNE (Fig.~\ref{fig:tsne}). Without domain adversarial training, CT and MRI features form clearly separated clusters, indicating a strong modality-induced domain shift. In contrast, features from different MRI sequences largely overlap, suggesting smaller intra-MRI variability. After introducing the domain discriminator, modality-specific clustering becomes less pronounced, indicating that modality-dependent cues are suppressed in the latent space. Although t-SNE is qualitative, these observations align with the improved robustness and cross-modality transfer observed in the quantitative results.

These findings should be interpreted within the scope of the study. First, the framework focuses on modality-level alignment (CT vs MRI) because experiments and feature analysis show this to be the dominant source of distribution shift, while finer-grained factors such as scanner or protocol variation are left for future work. Second, the subregion experiment intentionally reflects realistic annotation constraints, where part labels are available only in MRI; evaluating CT without CT subregion labels therefore directly tests cross-modality transfer under limited supervision. Third, the adversarial training schedule is empirically designed to ensure stable optimization in this multi-stage setting. Finally, the study is centered on pancreas segmentation, and extending the framework to other organs and tasks remains future work.
\label{sec:typestyle}
\section{Conclusion}
\label{sec:majhead}

In this study, we developed a unified pancreas segmentation framework that uses domain-adversarial learning to learn modality-invariant anatomical representations, enabling a single model to segment the pancreas across CT and MRI. Evaluation on 4,604 multi-center scans and multiple external OOD datasets demonstrated robust cross-modality performance and generalization. The learned representations also transferred effectively to downstream pancreas subregion segmentation using limited annotations from a single modality. In particular, the model achieved pancreatic head--body--tail segmentation on CT without using CT subregion annotations during downstream training, supporting a practical solution to fine-grained label scarcity across imaging modalities.

\bibliographystyle{IEEEtran}
\bibliography{refs}

\end{document}